\documentclass[10pt,twocolumn,letterpaper]{article}

\usepackage{iccv}
\usepackage{times}
\usepackage{epsfig}
\usepackage{graphicx}
\usepackage{amsmath}
\usepackage{amssymb}

\usepackage{tikz}
\usepackage{comment}
\usepackage{amsmath,amssymb} 
\usepackage{color}

\usepackage{multicol}
\usepackage{multirow}
\usepackage{booktabs}

\newlength\savewidth

\usepackage[breaklinks=true,bookmarks=false]{hyperref}

\iccvfinalcopy 

\def\iccvPaperID{0002} 

\ificcvfinal\fi

\begin{document}

\title{Multi-Modal Dataset Acquisition for Photometrically Challenging Objects}

\author{HyunJun Jung$^{\ast, 1}$
\and
Patrick Ruhkamp$^{\ast, 1,2}$
\and
Nassir Navab$^{1}$
\and
Benjamin Busam$^{1,2}$
\and
$^{\ast}$ equal contribution,$^{1}$ TU Munich, $^{2}$ 3Dwe.ai\\
{\tt\small hyunjun.jung, p.ruhkamp, b.busam@tum.de}
}

\maketitle
\ificcvfinal\thispagestyle{empty}\fi

\begin{abstract}
      This paper addresses the limitations of current datasets for 3D vision tasks in terms of accuracy, size, realism, and suitable imaging modalities for photometrically challenging objects. We propose a novel annotation and acquisition pipeline that enhances existing 3D perception and 6D object pose datasets. Our approach integrates robotic forward-kinematics, external infrared trackers, and improved calibration and annotation procedures. We present a multi-modal sensor rig, mounted on a robotic end-effector, and demonstrate how it is integrated into the creation of highly accurate datasets. Additionally, we introduce a freehand procedure for wider viewpoint coverage. Both approaches yield high-quality 3D data with accurate object and camera pose annotations. Our methods overcome the limitations of existing datasets and provide valuable resources for 3D vision research.
\end{abstract}
\section{Introduction}
\begin{figure}[!b]
      \centering
      \vspace{-3mm}
      \includegraphics[width=0.9\columnwidth]{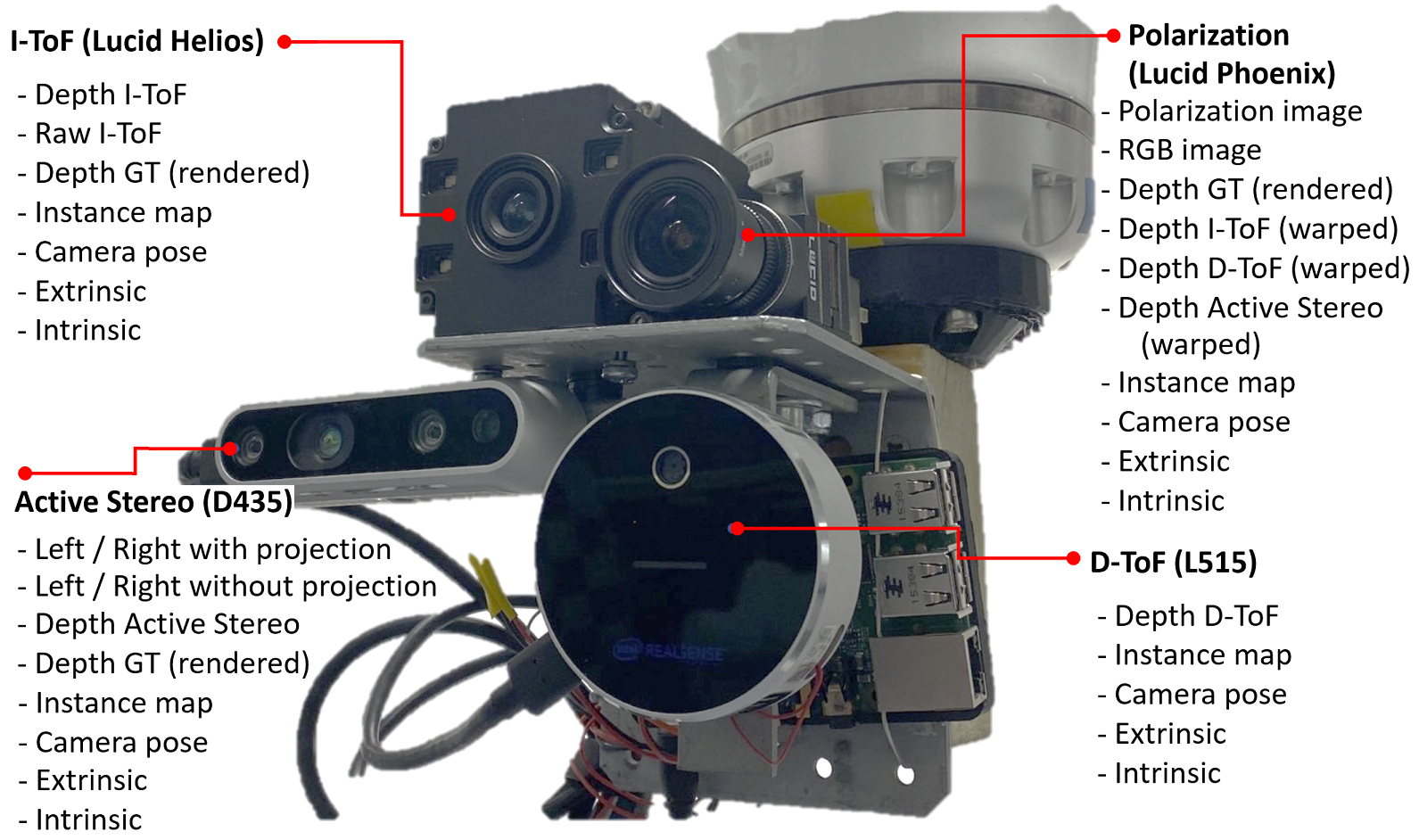}
      \vspace{1mm}
      \caption{Our camera rig comprises a custom multi-modal sensor setup, including Active Stereo, I-ToF, D-ToF and RGB-P sensors, that is triggered by a Raspberry Pi. The rig is mounted on a robot end-effector or infrared tracking marker.
      }
      \label{fig:sensor_rig}
\end{figure}

Are current datasets for 3D vision tasks meeting the requirements of accuracy, size, realism, and suitable imaging modalities, especially for photometrically challenging objects? For instance objects without texture, reflective surfaces or transparency - as is the case for common everyday objects like cups, glasses, or cutlery. This paper addresses these limitations by proposing a novel annotation and acquisition pipeline that enables the creation of highly accurate multi-modal (cf. Fig.~\ref{fig:sensor_rig}) datasets. By integrating robotic forward-kinematics, external infrared trackers, and improved calibration and annotation procedures, our approach has successfully enhanced recent 3D perception datasets~\cite{hammer} (depth, NeRF, sensor fusion) and instance- and category-level 6D object pose datasets~\cite{PhoCal}, which will be featured in the BOP challenge.
\begin{figure*}[!t]
 \centering
    \includegraphics[width=\linewidth]{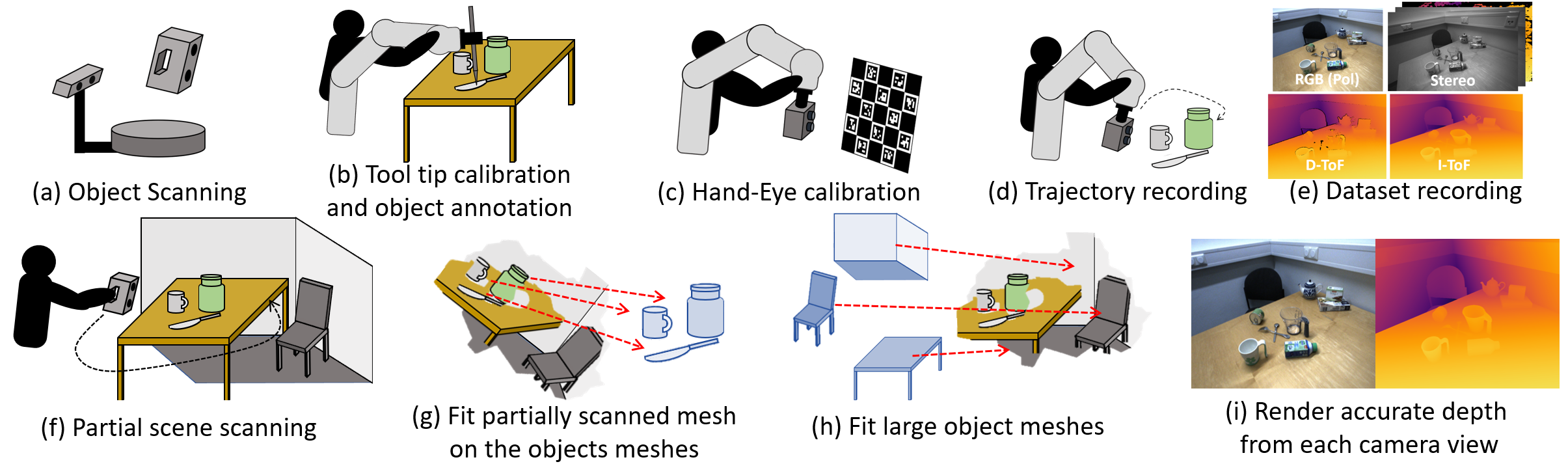}
    \vspace{-4mm}
    \caption{\textbf{Data Acquisition with Robotic Forward-Kinematics.} 
    We leverage highly accurate robot's kinematic to obtain the accurate 3D vision dataset. 
    (a)-(e) Annotation of the 6D pose of smaller objects close to the robot using the robot's highly accurate kinematics. (f)-(i) Annotation of background objects that are farther away using partial scanned mesh.
    }
    \label{fig:annotation_overview}
\end{figure*}
\section{Related Work}
\begin{figure*}[!t]
 \centering
    \includegraphics[width=\linewidth]{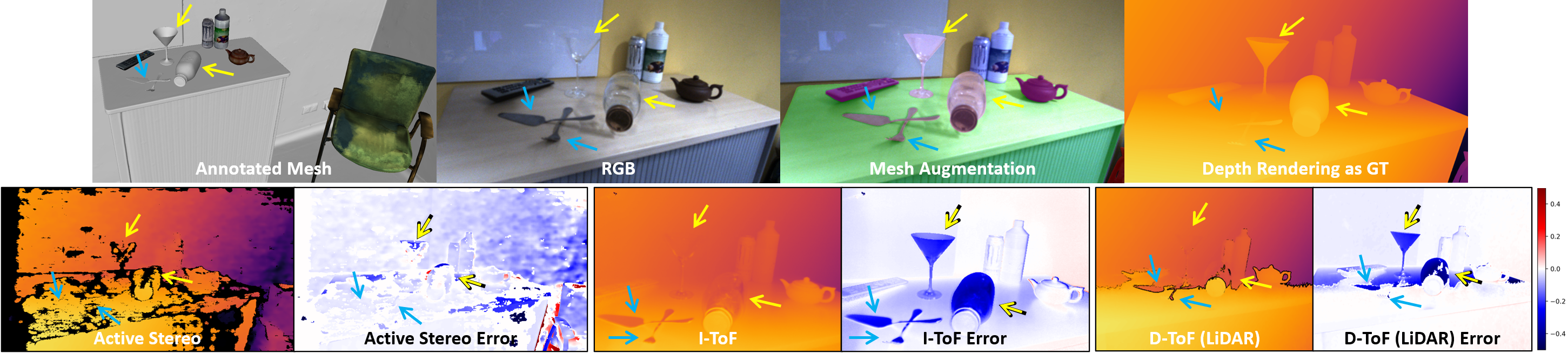}
    \vspace{-4mm}
    \caption{\textbf{Data Quality.} A complete annotation of the scene with mesh allows for accurate depth map rendering from any viewpoint. This serves as ground truth to analyze sensor errors for different scene structures. E.g., For ToF sensors, transparent objects such as glass (marked yellow) are undetectable and reflective objects (marked cyan) cause reflection induced error due to their measurement principle}
    \label{fig:dataset_quality}
    \vspace{-3mm}
\end{figure*}
The computer vision community relies on publicly available datasets to evaluate 3D vision tasks. In the context of depth estimation, early dataset~\cite{scharstein2002taxonomy} used passive multi-view stereo cameras, which suffered from limitations in textureless regions and restricted scenarios. Active sensor setups, such as active stereo and pattern projection sensors, addressed these issues by introducing artificial patterns and handled unconstrained scenarios~\cite{silberman2012indoor,sturm2012benchmark,aanaes2016IJCV,CroMo}. However, these sensors have their own artifacts, including bias, jitter~\cite{zhang2018activestereonet}, blurriness~\cite{jung2023importance}, and the inability to estimate depth on certain surfaces, necessitating post-processing via human annotation~\cite{silberman2012indoor,replica19arxiv}. Time of Flight (ToF) sensors, such as dToF-LiDAR and iToF, use light's traveling time for distance measurement and offer sharper depth measurements but can introduce artifacts like phase wrapping~\cite{jung2021wild} (iToF), multi-path interference (MPI)~\cite{mpi_2010,Fuchs2010MultipathIC,mpi_2012}, and material-dependent artifacts~\cite{jung2023importance}. Despite these artifacts, datasets created from these sensors are commonly used without evaluating depth quality.

In the domain of 6D pose estimation, various datasets have been developed. Commonly used datasets like LineMOD~\cite{hinterstoisser2011multimodal}, YCB~\cite{xiang2018posecnn} and NOCS~\cite{wang2019normalized} provide images with annotated object poses using checkerboards, RGBD cameras, or a combination of both. However, the annotation quality of these datasets is reported to be inaccurate due to limitations of checkerboard-based localization and errors from depth sensors~\cite{busam2020like,PhoCal,jung2023housecat6d}. More accurate pose annotation methods have been proposed, such as using multiview keypoints localized by checkerboards~\cite{liu2020keypose,liu2021stereobj}. These methods significantly improve annotation accuracy compared to depth-based annotation and even manage to annotate the photometrically challenging objects like glasses~\cite{liu2020keypose}. However, limitations in the acquisition pipeline, where a robotic arm is only used for scene scanning while object annotation relies on 2D keypoint annotation, result in considerable annotation errors (3.4mm~\cite{liu2020keypose}).

\begin{figure*}[t!]
 \centering
    \includegraphics[width=1.0\linewidth]{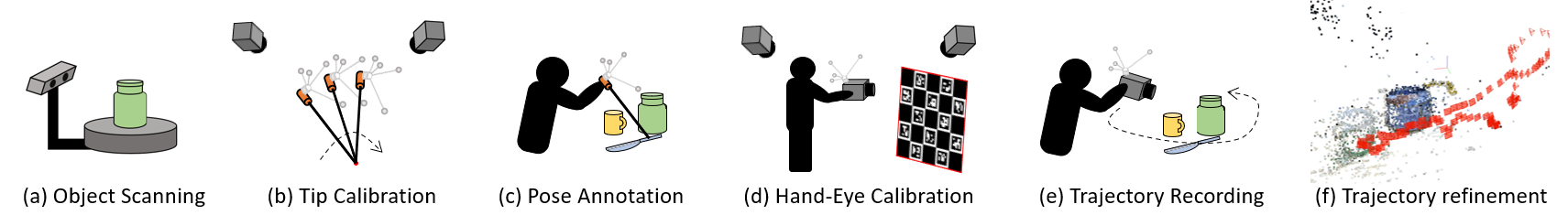}
    \vspace{-3mm}
    \caption{\textbf{Data Acquisition Pipeline for Free-hand Camera-Rig.} (a): Pre-scanning 3D models.
(b): Pivot calibration for measuring tip.
(c): Object pose annotation using the measuring tip.
(d): Hand-Eye-Calibration for camera center calibration.
(e): Camera trajectory recording.
(f): Post-processing to minimize synchronization errors.}
    \label{fig:pipeline}
\end{figure*}
\begin{figure*}[!t]
 \centering
    \vspace{-2mm}
    \includegraphics[width=1.0\linewidth]{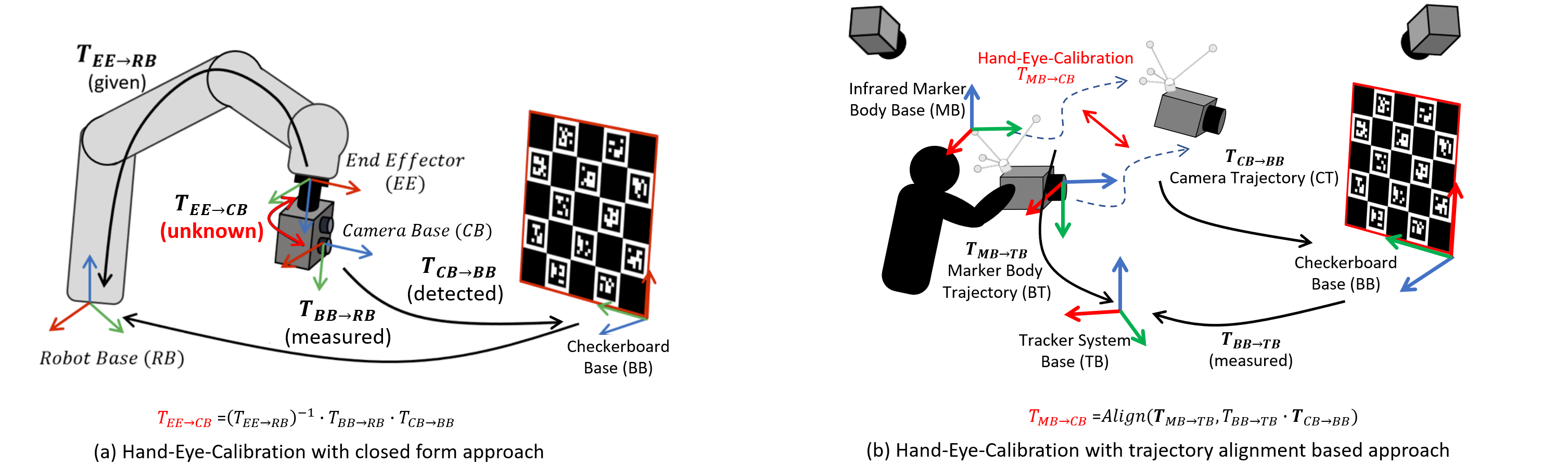}
    \caption{\textbf{Hand-Eye-Calibration.} We use two hand-eye calibration methods: (a) Closed-form approach for the robotic approach, and (b) Trajectory alignment approach for the external tracker approach that is more robust against error.}
    \label{fig:handeye_calibration}
\end{figure*}

\section{Dataset Acquisition}
\label{sec:dataset}
To overcome the limitations of existing datasets, we propose new paradigms for acquiring high-quality and multi-modal datasets for 3D vision tasks. Our approach involves a unique multi-modal sensor rig that incorporates various depth modalities (Fig.~\ref{fig:sensor_rig}). We leverage this setup to accurately measure the surfaces of objects and annotate both objects and scenes, which particularly includes photometrically challenging objects.
We introduce a robotic setup that enables the annotation of 6D object poses and camera poses. By utilizing precise forward kinematics of the robotic arm, we achieve highly accurate annotations for both depth and 6D pose datasets, as illustrated in Fig.~\ref{fig:annotation_overview}.

While the robotic setup provides precise annotations, it has limitations in terms of its working range (maximum 800mm radius) and joint limitations, which affect the distribution of camera poses for the 6D pose dataset. To address this issue, we employ a freehand procedure that ensures accurate data recording using an infrared tracking system and post-processing techniques, including multiple calibrations and trajectory refinement (Fig.~\ref{fig:pipeline}). Our freehand method offers a wider viewpoint coverage and more accurate object pose annotation compared to existing datasets, while maintaining better overall annotation quality.

Both our proposed dataset annotation methods, robotic and freehand, follow the same underlying principle for acquiring high-quality 3D data. 1. object or scene scanning, 2. measuring 20-30 high-quality surface points of the objects using a tracked tool tip, 3. annotating the object pose using point correspondence followed by ICP, and 4. recording the scene with a tracked camera. Although they share the same principle, each step requires different calibration and post-processing procedures to ensure data quality.

\subsection{Robotic Approach}
In our robotic approach, we utilize the KUKA LBR iiwa 7 R800 robot, which has a position accuracy of $\pm0.1$ mm. To scan small objects, we use the EinScan-SP table-top scanner, and for larger objects, we employ the Artec Eva hand-held scanner. To handle challenging materials, we apply the AESUB Blue 3D scanning spray before scanning.
 
\noindent\textbf{Object Pose Annotation.}
After acquiring the object meshes, we attach a measuring tool tip to the robot's End-Effector (EE) and perform tool tip calibration. We then capture accurate surface points of the objects using the tool tip, which are later used to annotate the object poses from the robot's base using point correspondence followed by ICP with the scanned object meshes. The pose error for the object pose annotation step is measured as 0.20 mm (RMSE) and 0.38\textdegree~\cite{PhoCal}. The procedure is depicted in Fig.~\ref{fig:annotation_overview} (a)-(e).

For the depth dataset, we also annotate background objects such as walls and tables to render the entire scene (cf.~Fig.\ref{fig:dataset_quality}). However, the limited working range of the robot prevents us from annotating the background with the robotic arm. Thus, we first scan the scene with a hand-held scanner to obtain a partial mesh. We then align the partial scanned mesh with the objects annotated from the robot base. Next, we fit the background meshes to the partial scanned mesh to align them with the robot base. 
The extra background annotation step is shown in Fig.~\ref{fig:annotation_overview} (f)-(i).

\noindent\textbf{Camera Pose Annotation.} Once all the objects and background is annotated, the scene is recorded with the camera rig. In the robotic approach, we attach the camera rig on the robot EE and perform hand-eye-calibration on each sensor to obtain the transformation matrix from the EE. For this, we use the closed form solution as in Fig.~\ref{fig:handeye_calibration} (a). When the checkerboard's transformation \(T_{BB->RB}\) is obtained via robotic tip and checkerboard is detected by camera \(T_{CB->BB}\), hand-eye-calibration \(T_{MB->CB}\) can be calculated as matrix multiplication of \(T_{RB->EE}\) (from forward kinematic), \(T_{BB->RB}\) and \(T_{CB->BB}\). During capturing the trajectory, we stop the robot on each frame before triggering the cameras to ensure perfect sync between the cameras and the robot. We accumulated all errors on the calibration step and obtained average RMSE of 0.86mm. Both object and camera pose annotation combined, the error for the entire pipeline is measured as 0.8mm (\cite{PhoCal}, supplementary Sec.2).
\begin{figure*}[t!]
 \centering
    \includegraphics[width=1.0\linewidth]{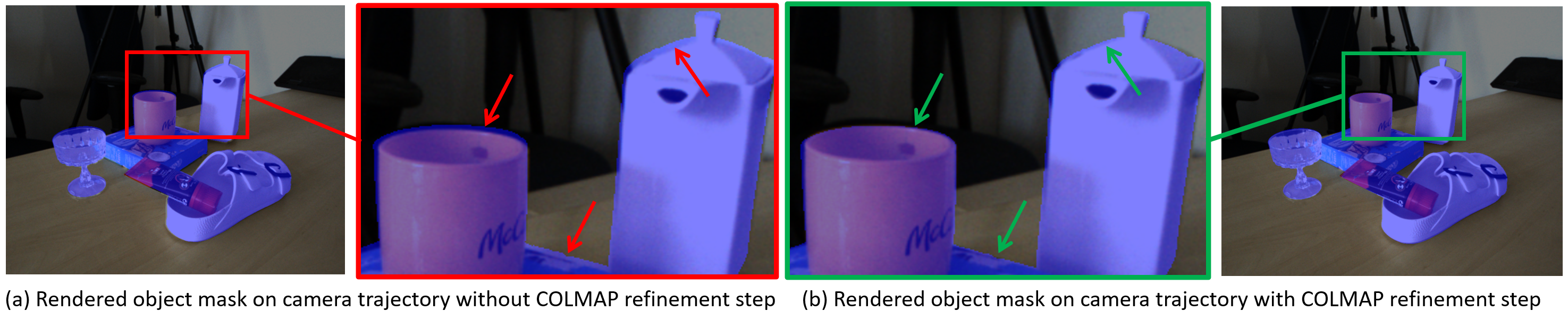}
    \vspace{-3mm}
    \caption{\textbf{Example of SfM based refinement.} SfM refinement can remove the subtle error that exist on camera's sudden movement}
    \label{fig:refinement}
\end{figure*}

\begin{figure*}[t!]
 \centering
    \vspace{-2mm}
    \includegraphics[width=1.0\linewidth]{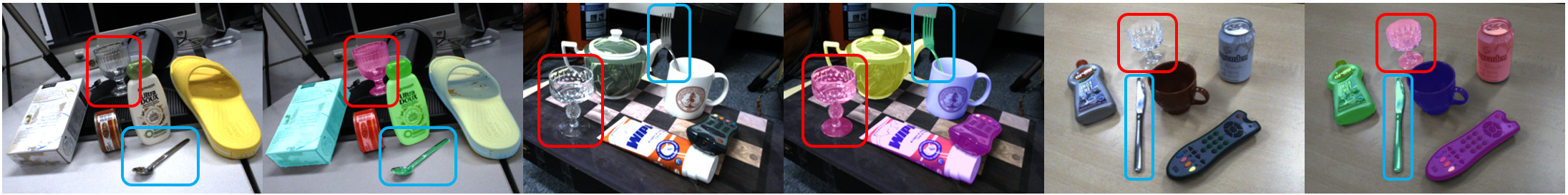}
    \vspace{-3mm}
    \caption{\textbf{Qualitative example of external tracker based annotation.} Red box highlights the annotation on glass object and Cyan box highlighligs the annotation on the reflective objects.}
    \label{fig:housecat_challenging_example}
    \vspace{-3mm}
\end{figure*}

\subsection{External Tracker Approach}
To overcome the limited pose coverage of the robotic arm-based annotation, we introduce a free-hand approach for acquiring high-quality 6D pose datasets. In this approach, we replace the robotic arm with an external tracking camera (ARTTRACK2, accuracy of 0.67 mm/0.12° in the static, 0.92 mm/0.16° in the dynamic case~\cite{jung2023housecat6d}) and achieve comparable accuracy to the robot-based annotation. The dataset acquisition pipeline is illustrated in Fig.~\ref{fig:pipeline}.

\noindent\textbf{Object Pose Annotation.} 
Object pose annotation in the free-hand approach follows a similar procedure to the robotic approach, with the exception of replacing the end-effector (EE) with an infrared (IR) tracking body that can be tracked by the external tracker. Similarly, we measure sparse surface points of the object using the calibrated tip and perform point correspondence with ICP to annotate the object from the tracker base. The tracker error in object annotation is measured as an average RMSE of $0.32$ mm in translation and $0.43^\circ$ in rotation, which is comparable to the robotic annotation method~\cite{jung2023housecat6d}.

\noindent\textbf{Camera Pose Annotation.} 
Replacing the EE with a tracking body on the camera and achieving high quality pose annotation introduces two challenges. Firstly, there is increased error in hand-eye calibration due to less precise hardware and more error propagation caused by the camera trajectory involving more rotations. Secondly, the accuracy of the tracking system is lower in the dynamic case due to synchronization issues.

To achieve a more reliable hand-eye calibration that is less susceptible to error, we utilize a trajectory of images instead of relying on a closed-form solution (cf. Fig.~\ref{fig:handeye_calibration}, (b)).
First, we obtain the checkerboard's pose with the calibrated tooltip \(T_{BB->TB}\). Then, the trajectory of the camera can be localized from the tracker system when the checkerboard is detected \(T_{CB->BB}\) via multiplication of \(T_{CB->BB}\) and \(T_{BB->TB}\). As camera's tracking marker trajectory is already localized from the tracker base \(T_{MB->TB}\), aligning the camera trajectory and the marker trajectory gives the offset pose that is the hand-eye-calibration matrix \(T_{MB->CB}\). We use the pose error between the aligned trajectory as the measurement of the error as 0.27 mm for translation and 0.42~\textdegree for rotation.

For camera synchronization, we initially achieve rough synchronization of all hardware using a hardware signal triggered by the tracker's pose input. Subsequently, we refine the time offset by applying ICP on the hardware trajectory. The refinement involves plotting the trajectory as a 2D distance graph with time on the x-axis and distance on the y-axis similar to~\cite{ar_sync1,ar_sync2}. Then, ICP is use to align the trajectory points and time offset is determined by the x-axis displacement. Despite these refinements, we observe slight pose offsets during sudden camera movements (cf. Fig.~\ref{fig:refinement} (a)). To address this issue, we employ SfM-based refinement techniques~\cite{colmap_1,colmap_2}, refining the camera trajectory using hand-selected fixed poses on a few frames. This refinement significantly improves the camera pose during sudden movements (Fig.~\ref{fig:refinement} (b)).
Quantifying the direct improvement achieved by SfM is challenging; hence, we evaluate the camera pose error using upper and lower bounds. The upper bound assumes dynamic errors caused by the tracking system, utilizing tracking errors from the dynamic case. In contrast, the lower bound assumes that dynamic errors are resolved, employing tracking errors from the static case. We evaluate our annotation quality by propagating the pose annotation error with the camera pose error, reporting an annotation quality range from 1.35 mm to 1.73 mm in terms of RMSE (cf.~\cite{jung2023housecat6d} (Sec. 3.5)). Fig.\ref{fig:housecat_challenging_example} shows the qualitative evaluation using object mask rendering that especially highlights the photometricall challenging objects.

\section{Conclusion}
\label{sec:discussion}
Our work proposes an annotation and acquisition pipeline that improves the accuracy and realism of 3D vision datasets. By integrating robotic forward-kinematics or external infrared trackers, together with enhanced calibration and annotation procedures, we provide valuable tools for dataset creation. 
The incorporation of external infrared trackers enhances camera pose viewpoint coverage, addressing limitations in existing datasets without the need for checkerboards to be present in the scene. 
Presented principles are particularly interesting for the creation of datasets with objects of high photometric complexity, such as glass or reflective and textureless materials (cf.~Figs.~\ref{fig:dataset_quality} and ~\ref{fig:housecat_challenging_example}).

\newpage

{\small
\bibliographystyle{ieee_fullname}
\bibliography{literature}
}

\end{document}